\def\year{2017}\relax
\documentclass[letterpaper]{article}
\usepackage{aaai17}
\usepackage{times}
\usepackage{helvet}
\usepackage{courier}
\usepackage{amsmath,amssymb,amsthm,bm}
\usepackage{algorithm}
\usepackage{algorithmic}
\usepackage{graphicx}
\usepackage{subfigure}
\usepackage{url}

\def\a{{\bf a}}
\def\b{{\bf b}}

\def\u{{\bf u}}
\def\v{{\bf v}}
\def\w{{\bf w}}
\def\x{{\bf x}}

\def\z{{\bf z}}

\def\A{{\bf A}}

\def\I{{\bf I}}

\def\0{{\bf 0}}
\def\1{{\bf 1}}
\def\2{{\bf 2}}
\def\3{{\bf 3}}
\def\4{{\bf 4}}
\def\5{{\bf 5}}
\def\6{{\bf 6}}
\def\7{{\bf 7}}
\def\8{{\bf 8}}
\def\9{{\bf 9}}

\def\DM{{\mathcal D}}

\def\SM{{\mathcal S}}

\def\EB{{\mathbb E}}

\def\RB{{\mathbb R}}

\newtheorem{theorem}{Theorem}
\newtheorem{lemma}{Lemma}

\newtheorem{assumption}{Assumption}
 
\begin{document}
%
\title{SCOPE: Scalable Composite Optimization for Learning on Spark}
\author{Shen-Yi Zhao, Ru Xiang, Ying-Hao Shi, Peng Gao \and Wu-Jun Li\\
National Key Laboratory for Novel Software Technology \\
Department of Computer Science and Technology, Nanjing University, China \\
\texttt{\{zhaosy,xiangr,shiyh,gaop\}@lamda.nju.edu.cn, liwujun@nju.edu.cn}
}
\maketitle
\begin{abstract}
Many machine learning models, such as logistic regression~(LR) and support vector machine~(SVM), can be formulated as composite optimization problems. Recently, many distributed stochastic optimization~(DSO) methods have been proposed to solve the large-scale composite optimization problems, which have shown better performance than traditional batch methods. However, most of these DSO methods might not be scalable enough. In this paper, we propose a novel DSO method, called \underline{s}calable \underline{c}omposite \underline{op}timization for l\underline{e}arning~({SCOPE}), and implement it on the fault-tolerant distributed platform \mbox{Spark}. SCOPE is both computation-efficient and communication-efficient. Theoretical analysis shows that SCOPE is convergent with linear convergence rate when the objective function is strongly convex.  Furthermore, empirical results on real datasets show that SCOPE can outperform other state-of-the-art distributed learning methods on Spark, including both batch learning methods and DSO methods.
\end{abstract}

\section{Introduction}
Many machine learning models can be formulated as composite optimization problems which have the following form with finite sum of some functions: $\underset{\w \in \RB^d} {\min}~P(\w) = \frac{1}{n}\sum_{i}^n f_{i}(\w)$, where $\w$ is the parameter to learn~(optimize), $n$ is the number of training instances, and $f_i(\w)$ is the loss function on the training instance $i$. For example,  $f_i(\w) = \log (1 + e^{-y_i \x_i^T \w})+\frac{\lambda}{2}\|\w\|^2$ in logistic regression~(LR), and $f_i(\w) = \max\{0, 1 - y_i \x_i^T \w\}+\frac{\lambda}{2}\|\w\|^2$ in support vector machine~(SVM), where $\lambda$ is the regularization hyper-parameter and $(\x_i, y_i)$ is the training instance $i$ with $\x_i \in \RB^d$ being the feature vector and $y_i\in \{+1,-1\}$ being the class label. Other cases like matrix factorization and deep neural networks can also be written as similar forms of composite optimization.

Due to its efficiency and effectiveness, stochastic optimization~(SO) has recently attracted much attention to solve the composite optimization problems in machine learning~\cite{DBLP:conf/nips/Xiao09,bottou-2010,DBLP:journals/jmlr/DuchiHS11,DBLP:journals/corr/SchmidtRB13,DBLP:conf/nips/Johnson013,DBLP:conf/nips/ZhangMJ13,DBLP:journals/corr/SDCA12,DBLP:conf/icml/Shalev-Shwartz014,DBLP:conf/nips/LinLX14,DBLP:conf/nips/Nitanda14}. Existing SO methods can be divided into two categories. The first category is stochastic gradient descent~(SGD) and its variants, such as stochastic average gradient~(SAG)~\cite{DBLP:journals/corr/SchmidtRB13} and stochastic variance reduced gradient~(SVRG)~\cite{DBLP:conf/nips/Johnson013}, which try to perform optimization on the primal problem. The second category, such as stochastic dual coordinate ascent~(SDCA)~\cite{DBLP:journals/corr/SDCA12}, tries to perform optimization with the dual formulation. Many advanced SO methods, such as SVRG and SDCA, are more efficient than traditional batch learning methods in both theory and practice for large-scale learning problems.

Most traditional SO methods are sequential which means that the optimization procedure is not parallelly performed. However, with the increase of data scale, traditional sequential SO methods may not be efficient enough to handle large-scale datasets. Furthermore, in this big data era, many large-scale datasets are distributively stored on a cluster of multiple machines. Traditional sequential SO methods cannot be directly used for these kinds of distributed datasets. To handle large-scale composite optimization problems, researchers have recently proposed several parallel SO~(PSO) methods for multi-core systems and distributed SO~(DSO) methods for clusters of multiple machines.

PSO methods perform SO on a single machine with multi-cores~(multi-threads) and a shared memory. Typically, synchronous strategies with locks will be much slower than asynchronous ones. Hence, recent progress of PSO mainly focuses on designing asynchronous or lock-free optimization strategies~\cite{DBLP:conf/nips/RechtRWN11,DBLP:conf/icml/LiuWRBS14,DBLP:conf/icml/HsiehYD15,NIPS2015_5821,DBLP:conf/aaai/ZhaoL16}.

DSO methods perform SO on clusters of multiple machines. DSO can be used to handle extremely large problems which are beyond the processing capability of one single machine. In many real applications especially industrial applications, the datasets are typically distributively stored on clusters. Hence, DSO has recently become a hot research topic. Many DSO methods have been proposed, including distributed SGD methods from primal formulation and distributed dual formulation. Representative distributed SGD methods include PSGD~\cite{DBLP:conf/nips/ZinkevichWLS10}, BAVGM~\cite{DBLP:conf/nips/ZhangMJ12} and Splash~\cite{DBLP:journals/corr/ZhangJ15}. Representative distributed dual formulations include DisDCA~\cite{DBLP:conf/nips/Yang13}, CoCoA~\cite{DBLP:conf/nips/MartinVMJSTM14} and CoCoA+~\cite{DBLP:conf/icml/MaSJJRT15}. Many of these methods provide nice theoretical proof about convergence and promising empirical evaluations. However, most of these DSO methods might not be scalable enough.

In this paper, we propose a novel DSO method, called \underline{s}calable \underline{c}omposite \underline{op}timization for l\underline{e}arning~(\mbox{{SCOPE}}), and implement it on the fault-tolerant distributed platform \mbox{Spark}~\cite{DBLP:conf/hotcloud/ZahariaCFSS10}. SCOPE is both computation-efficient and communication-efficient. Empirical results on real datasets show that SCOPE can outperform other state-of-the-art distributed learning methods on Spark, including both batch learning methods and DSO methods, in terms of scalability.

Please note that some asynchronous methods or systems, such as Parameter Server~\cite{DBLP:conf/osdi/LiAPSAJLSS14}, Petuum~\cite{DBLP:conf/kdd/XingHDKWLZXKY15} and the methods in~\cite{DBLP:conf/icml/ZhangK14,DBLP:journals/corr/Zhang0K15}, have also been proposed for distributed learning with promising performance. But these methods or systems cannot be easily implemented on Spark with the MapReduce programming model which is actually a bulk synchronous parallel~(BSP) model. Hence, asynchronous methods are not the focus of this paper. We will leave the design of asynchronous version of SCOPE and the corresponding empirical comparison for future study.

\section{SCOPE}

\subsection{Framework of SCOPE}
\begin{figure}[htb]
\begin{center}
\includegraphics[width=2.6in]{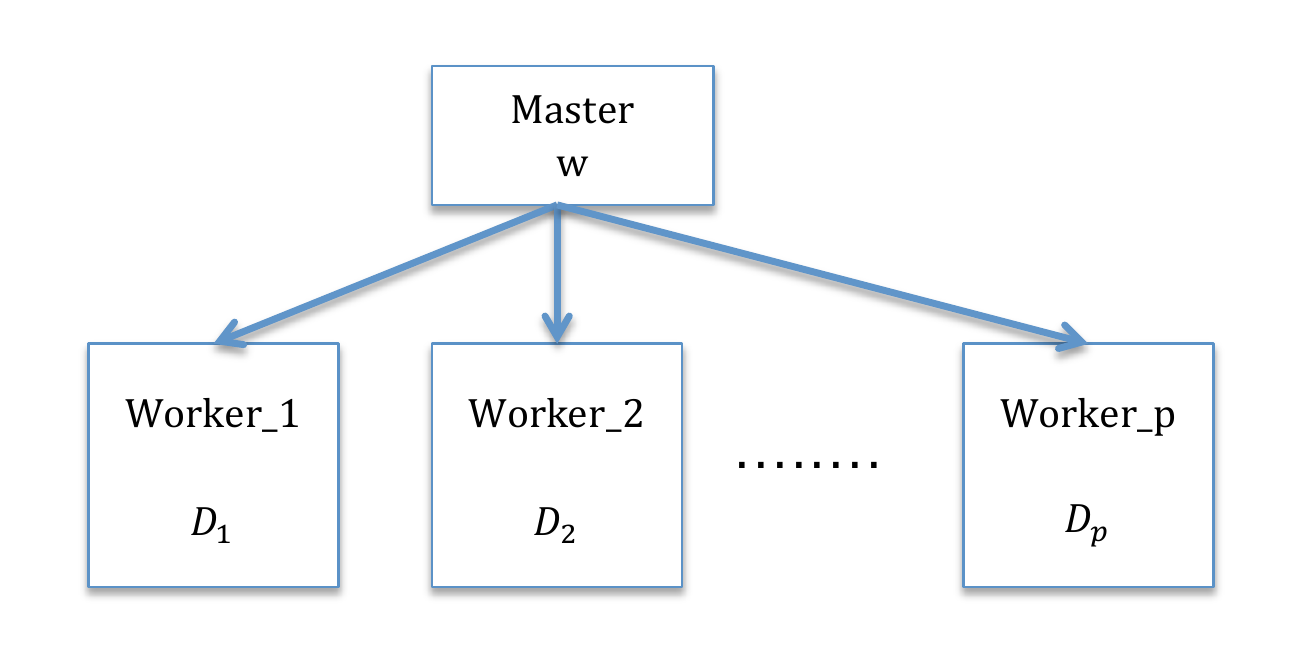}
\caption{\small Distributed framework of SCOPE.}
\label{fig:framework}
\end{center}
\end{figure}

SCOPE is based on a master-slave distributed framework, which is illustrated in Figure~\ref{fig:framework}. More specifically, there is a master machine~(called Master) and $p$~($p\geq 1$) slave machines~(called Workers) in the cluster. These Workers are called Worker$\_1$, Worker$\_2$, $\cdots$, and Worker$\_p$, respectively.

\subsubsection{Data Partition and Parameter Storage}
\begin{itemize}
\item For Workers: The whole dataset $\DM$ is distributively stored on all the Workers. More specifically, $\DM$ is partitioned into $p$ subsets, which are denoted as $\{\DM_1,\DM_2, \cdots, \DM_p\}$ with $\DM = \bigcup_{k=1}^p \DM_k$. $\DM_k$ is stored on Worker$\_k$. The data stored on different Workers are different from each other, which means that if $i \neq j$, $\DM_i \cap \DM_j = \emptyset$.
\item For Master: The parameter $\w$ is stored on the Master and the Master always keeps the newest version of $\w$.
\end{itemize}

Different Workers can not communicate with each other. This is similar to most existing distributed learning frameworks like MLlib~\cite{DBLP:journals/corr/MengMLlib15}, Splash, Parameter Server, and CoCoA and so on.

\subsubsection{Optimization Algorithm}
The whole optimization~(learning) algorithm is completed cooperatively by the Master and Workers:
\begin{itemize}
\item Task of Master: The operations completed by the Master are outlined in Algorithm~\ref{alg:mScope}. We can find that the Master has two main tasks. The first task is to compute the full gradient after all the \emph{local gradient sum} $\{\z_k\}$ have been received from all Workers, and then send the full gradient to all Workers. The second task is to update the parameter $\w$ after all the \emph{locally updated parameters} $\{\tilde{\u}_k\}$ have been received, and then send the updated parameter to all Workers. It is easy to see that the computation load of the Master is lightweight.

\item Task of Workers: The operations completed by the Workers are outlined in Algorithm~\ref{alg:wScope}. We can find that each Worker has two main tasks. The first task is to compute the sum of the gradients on its local data~(called \emph{local gradient sum}), i.e., $\z_k = \sum_{i \in \DM_k} \nabla f_i (\w)$ for Worker\_$k$, and then send the \emph{local gradient sum} to the Master. The second task is to train $\w$ by only using the local data, after which the Worker will send the \emph{locally updated parameters}, denoted as $\tilde{\u}_k$ for Worker\_$k$, to the Master and wait for the newest $\w$ from Master.
\end{itemize}

Here, $\w_t$ denotes the global parameter at the $t$th iteration and is stored on the Master. $\u_{k,m}$ denotes the local parameter at the $m$th iteration on Worker\_$k$.

\begin{algorithm}[!t]
\caption{Task of Master in SCOPE}
\label{alg:mScope}
\begin{algorithmic}
\STATE Initialization: $p$ Workers, $\w_0$;
\FOR{$t=0,1,2, \ldots,T$}
\STATE Send $\w_t$ to the Workers;
\STATE Wait until it receives $\z_1, \z_2, \ldots, \z_p$ from the $p$ Workers;
\STATE Compute the \emph{full gradient} $\z = \frac{1}{n} \sum_{k=1}^p \z_k$, and then send $\z$ to each Worker;
\STATE Wait until it receives $\tilde{\u}_1, \tilde{\u}_2, \ldots, \tilde{\u}_p$ from the $p$ Workers;
\STATE Compute $\w_{t+1} = \frac{1}{p} \sum_{k=1}^p \tilde{\u}_k$;
\ENDFOR
\end{algorithmic}
\end{algorithm}

\begin{algorithm}[!t]
\caption{Task of Workers in SCOPE}
\label{alg:wScope}
\begin{algorithmic}
\STATE Initialization: initialize $\eta$ and $c>0$;
\STATE For the Worker$\_k$:
\FOR{$t=0,1,2, \ldots,T$}
\STATE Wait until it gets the newest parameter $\w_t$ from the Master;
\STATE Let $\u_{k,0} = \w_t$, compute the \emph{local gradient sum} $\z_k = \sum_{i \in \DM_k} \nabla f_i(\w_t)$, and then send $\z_k$ to the Master;
\STATE Wait until it gets the full gradient $\z$ from the Master;
\FOR{$m=0$ to $M-1$}
\STATE Randomly pick up an instance with index $i_{k,m}$ from $\DM_k$;
\STATE $\u_{k, m+1} = \u_{k, m} - \eta (\nabla f_{i_{k,m}}(\u_{k,m}) - \nabla f_{i_{k,m}}(\w_t) + \z + c(\u_{k,m} - \w_t))$;
\ENDFOR
\STATE Send $\u_{k,M}$ or $\frac{1}{M}\sum_{m=1}^{M} \u_{k,m}$, which is called the \emph{locally updated parameter} and denoted as $\tilde{\u}_k$, to the Master;
\ENDFOR
\end{algorithmic}
\end{algorithm}

SCOPE is inspired by SVRG~\cite{DBLP:conf/nips/Johnson013} which tries to utilize full gradient to speed up the convergence of stochastic optimization. However, the original SVRG in~\cite{DBLP:conf/nips/Johnson013} is sequential. To design a distributed SVRG method, one natural strategy is to adapt the mini-batch SVRG~\cite{NIPS2014_5614} to distributed settings, which is a typical strategy in most distributed SGD frameworks like Parameter Server~\cite{DBLP:conf/osdi/LiAPSAJLSS14} and Petuum~\cite{DBLP:conf/kdd/XingHDKWLZXKY15}. In appendix\footnote{All the appendices and proofs of this paper can be found in the arXiv version of this paper~\cite{DBLP:journals/corr/ZhaoXSGL16}.}, we briefly outline the sequential SVRG and the mini-batch based distributed SVRG~(called DisSVRG). We can find that there exist three major differences between SCOPE and SVRG~(or DisSVRG).

The first difference is that in SCOPE each Worker \emph{locally} performs stochastic optimization by only using its native data~(refer to the update on $\u_{k,m+1}$ for each Worker\_$k$ in Algorithm~\ref{alg:wScope}). On the contrary, SVRG or DisSVRG perform stochastic optimization on the Master~(refer to the update on $\u_{m+1}$) based on the whole dataset, which means that we need to randomly pick up an instance or a mini-batch from the whole dataset $\DM$ in each iteration of stochastic optimization. The \emph{locally stochastic optimization} in SCOPE can dramatically reduce the communication cost, compared with DisSVRG with mini-batch strategy.

The second difference is the update rule of $\w_{t+1}$ in the Master. There are no \emph{locally updated parameters} in DisSVRG with mini-batch strategy, and hence the update rule of $\w_{t+1}$ in the Master for DisSVRG can not be written in the form of Algorithm~\ref{alg:mScope}, i.e., $\w_{t+1} = \frac{1}{p} \sum_{k=1}^p \tilde{\u}_k$.

The third difference is the update rule for $\u_{k,m+1}$ in \mbox{SCOPE} and $\u_{m+1}$ in SVRG or DisSVRG. Compared to SVRG, SCOPE has an extra term $c(\u_{k,m} - \w_t)$ in Algorithm~\ref{alg:wScope} to guarantee convergence, where $c >0$ is a parameter related to the objective function. The strictly theoretical proof will be provided in the following section about convergence. Here, we just give some intuition about the extra term $c(\u_{k,m} - \w_t)$. Since SCOPE puts no constraints about how to partition training data on different Workers, the data distributions on different Workers may be totally different from each other. That means the local gradient in each Worker can not necessarily approximate the full gradient. Hence, the term $\nabla f_{i_{k,m}}(\u_{k,m}) - \nabla f_{i_{k,m}}(\w_t) + \z$ is a bias estimation of the full gradient. This is different from SVRG whose stochastic gradient is an unbias estimation of the full gradient. The bias estimation $\nabla f_{i_{k,m}}(\u_{k,m}) - \nabla f_{i_{k,m}}(\w_t) + \z$ in SCOPE may lead $\u_{k,m+1}$ to be far away from the optimal value $\w^*$. To avoid this, we use the technique in the proximal stochastic gradient that adds an extra term $c(\u_{k,m} - \w_t)$ to make $\u_{k,m+1}$ not be far away from $\w_t$. If $\w_t$ is close to $\w^*$, $\u_{k,m+1}$ will also be close to $\w^*$. So the extra term in SCOPE is reasonable for convergence guarantee.  At the same time, it does not bring extra computation since the update rule in SCOPE can be rewritten as
\begin{align}
\u_{k, m+1} = &(1-c\eta)\u_{k, m} \nonumber \\
              &- \eta (\nabla f_{i_{k,m}}(\u_{k,m}) - \nabla f_{i_{k,m}}(\w_t) + \hat{\z}), \nonumber
\end{align}
where $\hat{\z} = \z - c \w_t$ can be pre-computed and fixed as a constant for different $m$.

Besides the above mini-batch based strategy~(DisSVRG) for distributed SVRG, there also exist some other distributed SVRG methods, including DSVRG~\cite{DBLP:journals/corr/LeeLMY2016}, KroMagnon~\cite{DBLP:journals/corr/ManiaPPRRJ15}, SVRGfoR~\cite{DBLP:journals/corr/KonecnyMR15} and the distributed SVRG in~\cite{DBLP:conf/icdm/DeG16}. DSVRG needs communication between Workers, and hence it cannot be directly implemented on Spark. KroMagnon focuses on asynchronous strategy, which cannot be implemented on Spark either. SVRGfoR can be implemented on Spark, but it provides no theoretical results about the convergence. Furthermore, SVRGfoR is proposed for cases with unbalanced data partitions and sparse features. On the contrary, our SCOPE can be used for any kind of features with theoretical guarantee of convergence. Moreover, in our experiment, we find that our SCOPE can outperform SVRGfoR. The distributed SVRG in~\cite{DBLP:conf/icdm/DeG16} cannot be guaranteed to converge because it is similar to the version of SCOPE with $c =0$.

EASGD~\cite{DBLP:conf/nips/ZhangCL15} also adopts a parameter like $c$ to control the difference between the local update and global update. However, EASGD assumes that each worker has access to the entire dataset while SCOPE only requires that each worker has access to a subset. Local learning strategy is also adopted in other problems like probabilistic logic programs~\cite{DBLP:conf/ecai/RiguzziBZCL16}.

\subsection{Communication Cost}\label{sec:communication}
Traditional mini-batch based distributed SGD methods, such as DisSVRG in the appendix, need to transfer parameter $\w$ and stochastic gradients frequently between Workers and Master. For example, the number of communication times
is $O(TM)$ for DisSVRG. Other traditional mini-batch based distributed SGD methods have the same number of communication times. Typically, $M = \Theta(n)$. Hence, traditional mini-batch based methods have $O(Tn)$ number of
communication times, which may lead to high communication cost.


Most training~(computation) load of SCOPE comes from the inner loop of Algorithm~\ref{alg:wScope}, which is done at local Worker without any communication. It is easy to find that the number of communication times in SCOPE is $O(T)$,
which is dramatically less than $O(Tn)$ of traditional mini-batch based distributed SGD or distributed SVRG methods. In the following section, we will prove that SCOPE has a linear convergence rate in terms of the iteration number
$T$. It means that to achieve an $\epsilon$-optimal solution\footnote{$\hat{\w}$ is called an $\epsilon$-optimal solution if $\EB \|\hat{\w} - \w^* \|^2\leq \epsilon$ where $\w^*$ is the optimal solution.}, $T=O(\log \frac{1}{\epsilon})$.
Hence, $T$ is typically not large for many problems. For example, in most of our experiments, we can achieve convergent results with $T\leq 10$. Hence, SCOPE is communication-efficient. SCOPE is a synchronous framework, which means
that some waiting time is also needed for synchronization. Because the number of synchronization is also $O(T)$, and $T$ is typically a small number. Hence, the waiting time is also small.

\subsection{SCOPE on Spark}

One interesting thing is that the computing framework of SCOPE is quite suitable for the popular distributed platform Spark. The programming model underlying Spark is MapReduce, which is actually a BSP model. In SCOPE, the task of Workers that computes \emph{local gradient sum} $\z_k$ and the training procedure in the inner loop of Algorithm~\ref{alg:wScope} can be seen as the Map process since both of them only use local data. The task of Master that computes the average for both \emph{full gradient} $\z$ and $\w_{t+1}$ can be seen as the Reduce process.

The MapReduce programming model is essentially a synchronous model, which need some synchronization cost. Fortunately, the number of synchronization times is very small as stated above. Hence, both communication cost and waiting time are very small for SCOPE. In this paper, we implement our SCOPE on Spark since Spark has been widely adopted in industry for big data applications, and our SCOPE can be easily integrated into the data processing pipeline of those organizations using Spark.

\section{Convergence of SCOPE}\label{sec:convergence}
In this section, we will prove the convergence of SCOPE when the objective functions are strongly convex. We only list some Lemmas and Theorems, the detailed proof of which can be found in the appendices~\cite{DBLP:journals/corr/ZhaoXSGL16}.

For convenience, we use $\w^*$ to denote the optimal solution. $\|\cdot\|$ denotes the $L_2$ norm $\| \cdot \|_2$.
We assume that $n = pq$, which means that each Worker has the same number of training instances and $| \DM_1 | = | \DM_2 | = \cdots = | \DM_p | = q$. In practice, we can not necessarily guarantee that these $| \DM_k |$s are the same. However, it is easy to guarantee that $\forall i,j, | (|\DM_i| - |\DM_j|) | \leq 1$, which will not affect the performance.

We define $p$ local functions as $F_k(\w) = \frac{1}{q} \sum_{i \in \DM_k} f_i(\w)$,
where $k = 1, 2, \ldots, p$. Then we have $P(\w) = \frac{1}{p} \sum_{k=1}^p F_k(\w)$.

To prove the convergence of SCOPE, we first give two assumptions which have also been widely adopted by most existing stochastic optimization algorithms for convergence proof.
\begin{assumption}[Smooth Gradient]\label{ass:smooth}
There exists a constant $L>0$ such that $\forall \a, \b \in \RB^d$ and $i = 1, 2, \ldots, n$, we have $\| \nabla f_i(\a) - \nabla f_i(\b) \| \leq L \| \a - \b \|$.
\end{assumption}

\begin{assumption}[Strongly Convex]\label{ass:stroconv}
For each local function $F_k(\cdot)$, there exists a constant $\mu > 0$ such that $\forall \a, \b \in \RB^d$, we have $F_k(\a) \geq F_k(\b) + \nabla F_k(\b)^T(\a - \b) + \frac{\mu}{2} \| \a - \b \|^2$.
\end{assumption}
Please note that these assumptions are weaker than those in~\cite{DBLP:journals/corr/ZhangJ15,DBLP:conf/icml/MaSJJRT15,DBLP:conf/nips/MartinVMJSTM14}, since we do not need each $f_i(\w)$ to be convex and we do not make any assumption about the Hessian matrices either.


\begin{lemma}\label{lemma:gamma_m}
Let $\gamma_m = \frac{1}{p} \sum_{k=1}^p\EB \| \u_{k,m} - \w^* \|^2$. If $c > L - \mu$, then we have $\gamma_{m+1} \leq [1- \eta(2\mu + c)]\gamma_m + (c\eta + 3L^2 \eta^2)\gamma_0$.
\end{lemma}

Let $\alpha = 1 - \eta(2\mu + c)$, $\beta = c\eta + 3L^2 \eta^2$. Given $L$ and $\mu$ which are determined by the objective function, we can always guarantee $0<\alpha < 1$, $0<\beta < 1$, and $\alpha + \beta < 1$ by setting $\eta <\min\{\frac{2\mu}{3L^2},\frac{1}{2\mu+c}\}$.
We have the following theorems:
\begin{theorem}\label{cor1}
If we take $\w_{t+1}=\frac{1}{p}\sum_{k=1}^p \u_{k,M}$, then we can get the following convergence result:
\begin{equation}
\EB \| \w_{t+1} - \w^* \|^2 \leq (\alpha^M + \frac{\beta}{1 - \alpha})\EB \| \w_t - \w^* \|^2 .\nonumber
\end{equation}
\end{theorem}
When $M > \log_{\alpha}^{\frac{1-\alpha-\beta}{1-\alpha}}$, $\alpha^M + \frac{\beta}{1 - \alpha} <1$, which means we can get a linear convergence rate if we take $\w_{t+1}=\frac{1}{p}\sum_{k=1}^p \u_{k,M}$.

\begin{theorem}\label{cor2}
If we take $\w_{t+1}=\frac{1}{p}\sum_{k=1}^p \tilde{\u}_k$ with $\tilde{\u}_k = \frac{1}{M}\sum_{m=1}^{M} \u_{k,m}$, then we can get the following convergence result:
\begin{equation}
\EB \| \w_{t+1} - \w^* \|^2 \leq (\frac{1}{M(1-\alpha)} + \frac{\beta}{1-\alpha})\EB \| \w_t - \w^* \|^2 .\nonumber
\end{equation}
\end{theorem}
When $M > \frac{1}{1-\alpha-\beta}$, $\frac{1}{M(1-\alpha)} + \frac{\beta}{1-\alpha} < 1$, which means we can also get a linear convergence rate if we take $\w_{t+1}=\frac{1}{p}\sum_{k=1}^p \tilde{\u}_k$ with $\tilde{\u}_k = \frac{1}{M}\sum_{m=1}^{M} \u_{k,m}$.

According to Theorem~\ref{cor1} and Theorem~\ref{cor2}, we can find that SCOPE gets a linear convergence rate when $M$ is larger than some threshold. To achieve an $\epsilon$-optimal solution, the computation complexity of each worker is $O((\frac{n}{p}+M)\log\frac{1}{\epsilon})$. In our experiment, we find that good performance can be achieved with $M = \frac{n}{p}$. Hence, SCOPE is computation-efficient.

\section{Impact of Parameter $c$}
In Algorithm~\ref{alg:wScope}, we need the parameter $c$ to guarantee the convergence of SCOPE. Specifically, we need $c > L-\mu$ according to Lemma~\ref{lemma:gamma_m}. Here, we discuss the necessity of $c$.

We first assume $c = 0$, and try to find whether Algorithm~\ref{alg:wScope} will converge or not. It means that in the following derivation, we always assume $c = 0$.

Let us define another local function:
\begin{align}
  F_k^{(t)}(\w) = F_k(\w) + (\z - \nabla F_k(\w_t))^T(\w-\w^*) \nonumber
\end{align}
and denote $\w_{k,t}^* = \underset{\w}{\arg\min}~F_k^{(t)}(\w).$

Let $\v_{k,m}= \nabla f_{i_{k,m}}(\u_{k,m}) - \nabla f_{i_{k,m}}(\w_t) + \z + c(\u_{k,m} - \w_t)$. When $c = 0$, $\v_{k,m}= \nabla f_{i_{k,m}}(\u_{k,m}) - \nabla f_{i_{k,m}}(\w_t) + \z$. Then, we have $\EB [\v_{k,m}|\u_{k,m}] = \nabla F_k^{(t)}(\u_{k,m})$ and $\nabla F_k^{(t)}(\w_t) = \z$. Hence, we can find that each local Worker actually tries to optimize the local function $F_k^{(t)}(\w)$ with SVRG based on the local data $\DM_k$. It means that if we set a relatively small $\eta$ and a relatively large $M$, the $\u_{k,m}$ will converge to $\w_{k,t}^*$.

Since $F_k^{(t)}(\w)$ is strongly convex, we have $\nabla F_k^{(t)}(\w_{k,t}^*) = 0$. Then, we can get
\begin{align*}
  \nabla F_k(\w_{k,t}^*) - \nabla F_k(\w^*) = \nabla F_k(\w_t) -\nabla F_k(\w^*) - \z.
\end{align*}
For the left-hand side, we have
\begin{align}
  \nabla F_k(\w_{k,t}^*) - \nabla F_k(\w^*) \approx \nabla^2 F_k(\w^*)(\w_{k,t}^* - \w^*). \nonumber
\end{align}
For the right-hand side, we have
\begin{align}
          &\nabla F_k(\w_t) -\nabla F_k(\w^*) - \z \nonumber \\
  =       &\nabla F_k(\w_t) -\nabla F_k(\w^*) - (\z - \nabla P(\w^*)) \nonumber \\
  \approx &\nabla^2 F_k(\w^*)(\w_t - \w^*) - \nabla^2 P(\w^*)(\w_t - \w^*). \nonumber
\end{align}
Combining the two approximations, we can get
\begin{align}
  \w_{k,t}^* - \w^* \approx (\I - \A_k^{-1}\A)(\w_t - \w^*), \nonumber
\end{align}
where $\A_k = \nabla^2 F_k(\w^*)$ and $\A = \nabla^2 P(\w^*)$ are two Hessian matrices for the local function
$F_k(\w^*)$ and the global function $P(\w^*)$, respectively. Assuming in each iteration we can always get the local optimal values for all local functions,
we have
\begin{align}\label{approx:iter}
  \w_{t+1} - \w^* \approx (\I - \frac{1}{p}\sum_{k=1}^p \A_k^{-1}\A)(\w_t - \w^*).
\end{align}

Please note that all the above derivations assume that $c=0$. From~(\ref{approx:iter}), we can find that Algorithm~\ref{alg:wScope} will not necessarily converge if $c=0$, and
the convergence property is dependent on the Hessian matrices of the local functions.

Here, we give a simple example for illustration. We set $n=p=2$ and $F_1(\w) = (\w-1)^2, F_2(\w) = 100(\w-10)^2$. We set a small step-size $\eta = 10^{-5}$ and a large $M = 4000$. The convergence results of SCOPE with different $c$ are presented in Table~\ref{Tab:c}.
\begin{table}[!thb]
\small
  \caption{Impact of $c$.}\label{Tab:c}
  \centering
  \begin{tabular}{|c|c|c|c|c|}
  \hline
    $c$ & 0 & 1 & 5 & 10 \\ \hline
    Converge? & No & No & No &Yes \\ \hline
   \end{tabular}
\end{table}

\subsection{Separating Data Uniformly}
If we separate data uniformly, which means that the local data distribution on each Worker is similar to the global
data distribution, then we have $\A_k \approx \A$ and $\| \I - \frac{1}{p}\sum_{i=1}^p \A_k^{-1}\A \| \approx 0$.
From~(\ref{approx:iter}), we can find that $c=0$ can make SCOPE converge for this special case.

\section{Experiment}
We choose logistic regression~(LR) with a $L_2$-norm regularization term to evaluate SCOPE and baselines. Hence, $P(\w)$ is defined as $P(\w) = \frac{1}{n} \sum_{i=1}^n \left[ \log(1 + e^{-y_i \x_i^T \w}) + \frac{\lambda}{2} \| \w \|^2 \right]$. The code can be downloaded from~\url{https://github.com/LIBBLE/LIBBLE-Spark/}.

\subsection{Dataset}
We use four datasets for evaluation. They are MNIST-8M, epsilon, KDD12 and Data-A. The first two datasets can be downloaded from the LibSVM website\footnote{https://www.csie.ntu.edu.tw/$\sim$cjlin/libsvmtools/datasets/}. MNIST-8M contains 8,100,000 handwritten digits. We set the instances of digits 5 to 9 as positive, and set the instances of digits 0 to 4 as negative. KDD12 is the dataset of Track 1 for KDD Cup 2012, which can be downloaded from the KDD Cup website\footnote{http://www.kddcup2012.org/}.
Data-A is a dataset from a data mining competition\footnote{http://www.yiban.cn/project/2015ccf/comp\_detail.php?cid=231}. The information about these datasets is summarized in Table~\ref{Tab:data}. All the data is normalized before training. The regularization hyper-parameter $\lambda$ is set to $10^{-4}$ for the first three datasets which are relatively small, and is set to $10^{-6}$ for the largest dataset Data-A. Similar phenomenon can be observed for other $\lambda$, which is omitted due to space limitation. For all datasets, we set $c = \lambda \times 10^{-2}$.

\begin{table}[ht]
\small
  \caption{Datasets for evaluation.}\label{Tab:data}
  \centering
  \begin{tabular}{|c|c|c|c|c|}
     \hline
     ~ & $\sharp$instances & $\sharp$features & memory &$\lambda$ \\ \hline
     MNIST-8M & 8,100,000 & 784 & 39G & 1e-4 \\ \hline
     epsilon & 400,000 & 2,000 & 11G & 1e-4 \\ \hline
     KDD12 & 73,209,277 & 1,427,495 & 21G & 1e-4 \\ \hline
     Data-A & 106,691,093 & 320 & 260G & 1e-6 \\
     \hline
   \end{tabular}
\end{table}

\subsection{Experimental Setting and Baseline}
\subsubsection{Distributed Platform}
We have a Spark cluster of 33 machines~(nodes) connected by 10GB Ethernet. Each machine has 12 Intel Xeon E5-2620 cores with 64GB memory. We construct two clusters, a small one and a large one, from the original 33 machines for our experiments. The small cluster contains $9$ machines, one master and eight slaves. We use 2 cores for each slave. The large cluster contains 33 machines, 1 master and 32 slaves. We use 4 cores for each slave. In both clusters, each machine has access to 64GB memory on the corresponding machine and one core corresponds to one Worker. Hence, the small cluster has one Master and 16 Workers, and the large cluster has one Master and 128 Workers. The small cluster is for experiments on the three relatively small datasets including MNIST-8M, epsilon and KDD12. The large cluster is for experiments on the largest dataset Data-A. We use Spark1.5.2 for our experiment, and implement our SCOPE in Scala.

\subsubsection{Baseline}
Because the focus of this paper is to design distributed learning methods for Spark, we compare SCOPE with distributed learning baselines which can be implemented on Spark. More specifically, we adopt the following baselines for comparison:
\begin{itemize}
\item MLlib\footnote{http://spark.apache.org/mllib/}~\cite{DBLP:journals/corr/MengMLlib15}: MLlib is an open source library for distributed machine learning on Spark. It is mainly based on two optimization methods: mini-batch based distributed SGD and distributed lbfgs. We find that the distributed SGD method is much slower than distributed lbfgs on Spark in our experiments. Hence, we only compare our method with distributed lbfgs for MLlib, which is a batch learning method.
\item LibLinear\footnote{https://www.csie.ntu.edu.tw/$\sim$ cjlin/liblinear/}~\cite{DBLP:conf/bigdataconf/LinTLL14}: LibLinear is a distributed Newton method, which is also a batch learning method.
\item Splash\footnote{http://zhangyuc.github.io/splash}~\cite{DBLP:journals/corr/ZhangJ15}: Splash is a distributed SGD method by using the local learning strategy to reduce communication cost~\cite{DBLP:conf/nips/ZhangMJ12}, which is different from the mini-batch based distributed SGD method.
\item CoCoA\footnote{https://github.com/gingsmith/cocoa}~\cite{DBLP:conf/nips/MartinVMJSTM14}: CoCoA is a distributed dual coordinate ascent method by using local learning strategy to reduce communication cost, which is formulated from the dual problem.
\item CoCoA+\footnote{https://github.com/gingsmith/cocoa}~\cite{DBLP:conf/icml/MaSJJRT15}: CoCoA+ is an improved version of CoCoA. Different from CoCoA which adopts average to combine local updates for global parameters, CoCoA+ adopts adding to combine local updates.
\end{itemize}

We can find that the above baselines include state-of-the-art distributed learning methods with different characteristics. All the authors of these methods have shared the source code of their methods to the public. We use the source code provided by the authors for our experiment. For all baselines, we try several parameter values to choose the best performance.

\subsection{Efficiency Comparison with Baselines}
\label{4.5}

\begin{figure}[!htb]
  \centering
  \subfigure[MNIST-8M]{\includegraphics[width=1.4in]{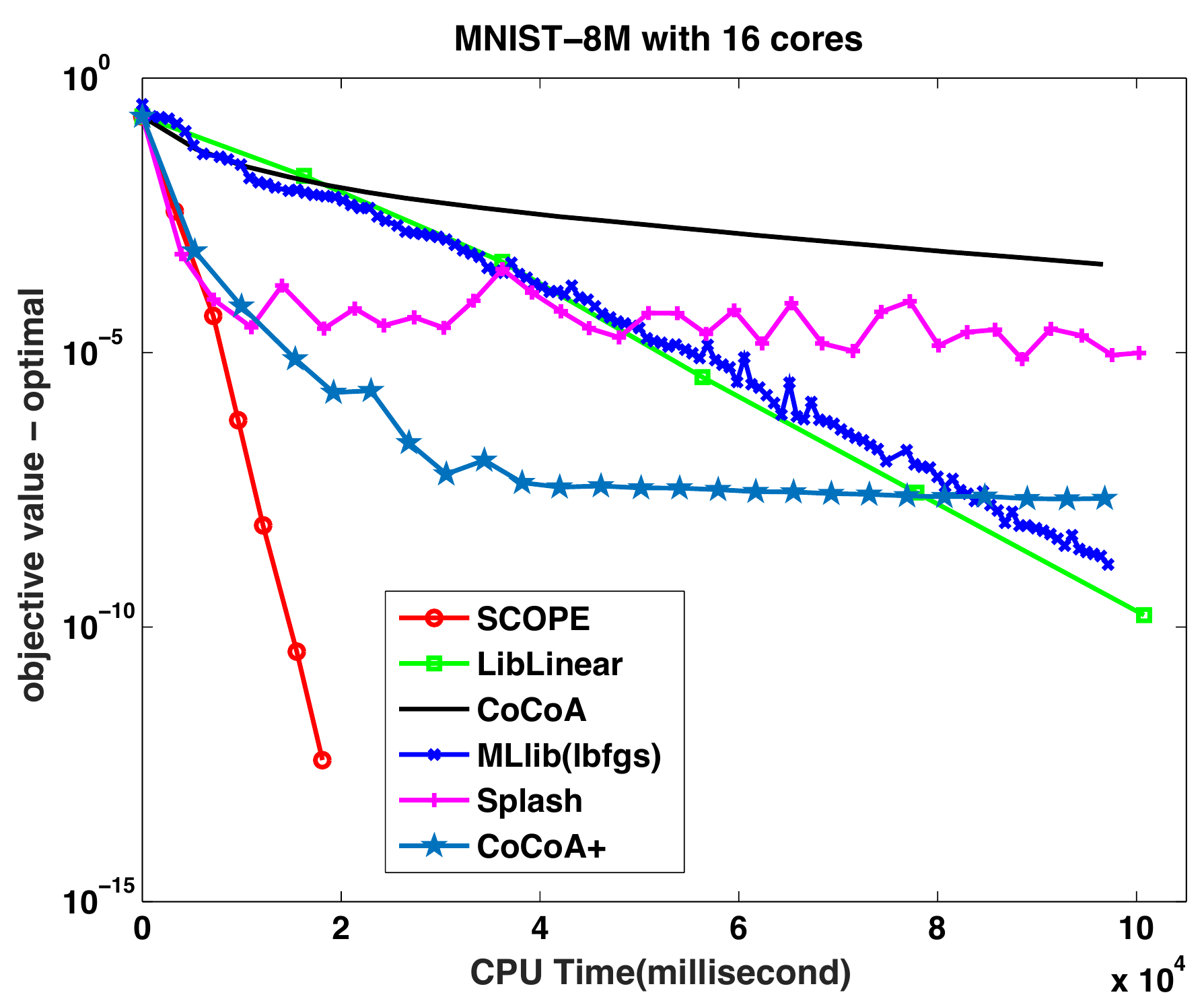}}
  \subfigure[epsilon]{\includegraphics[width=1.4in]{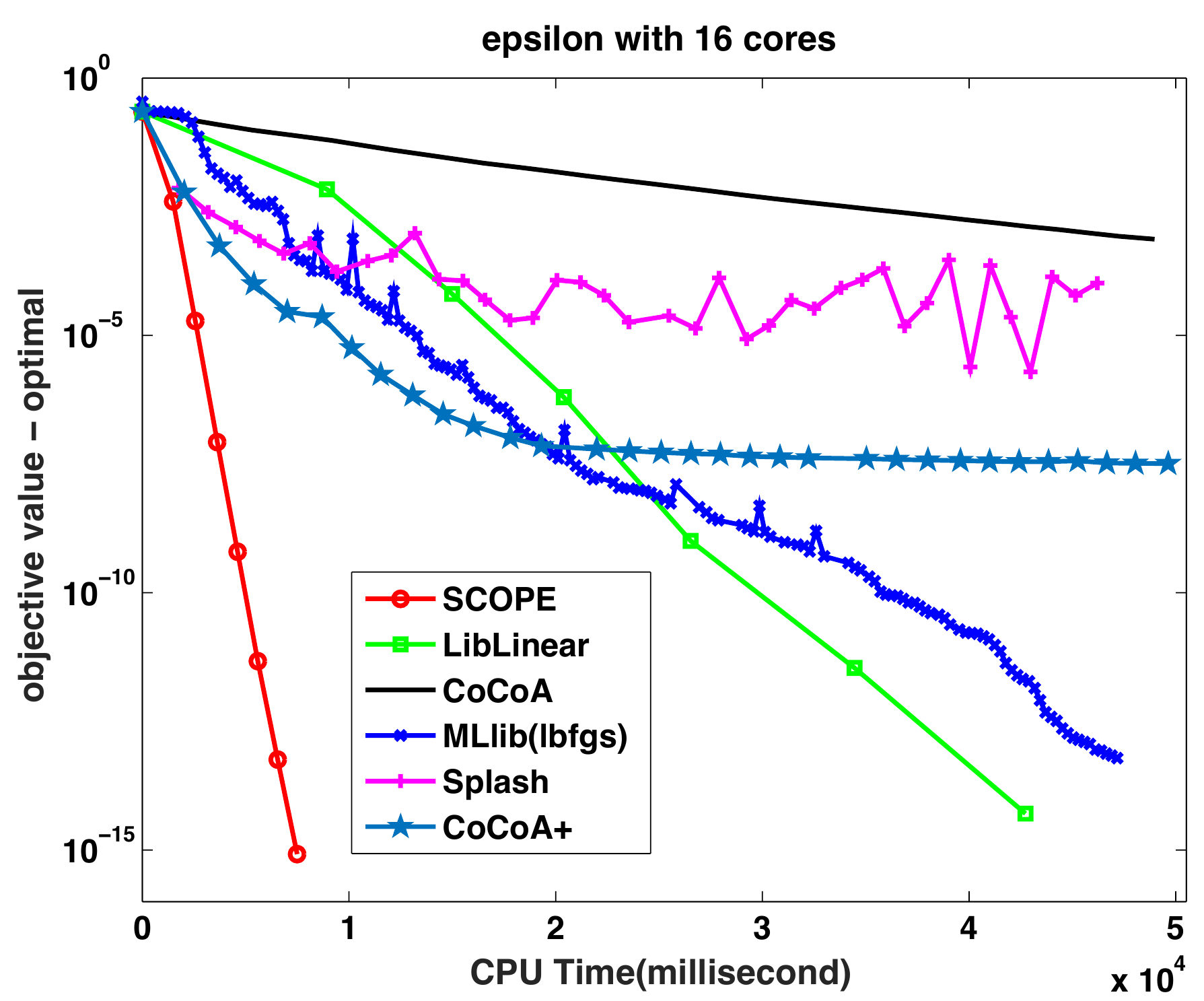}}
  \subfigure[KDD12]{\includegraphics[width=1.4in]{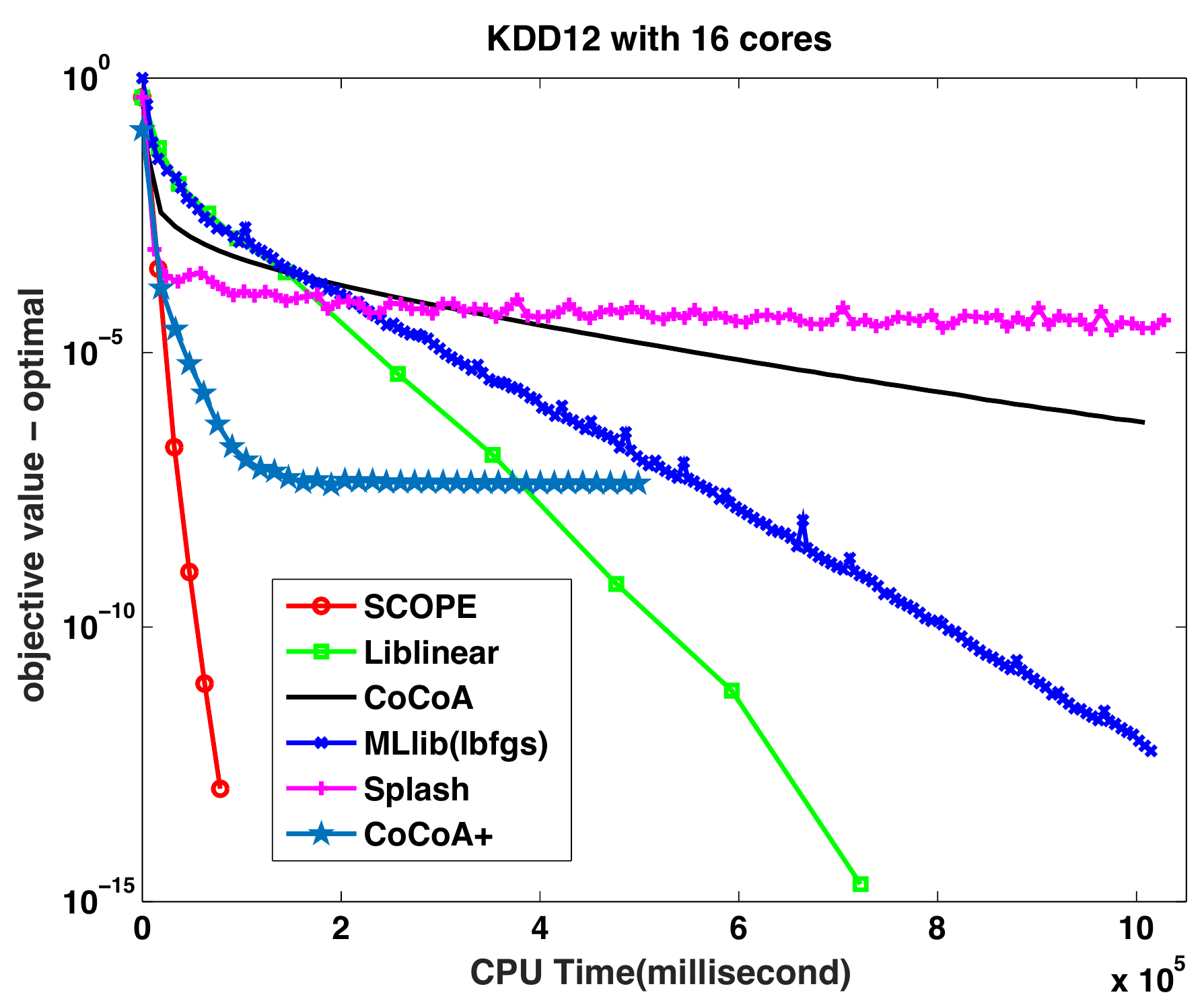}}
  \subfigure[Data-A]{\includegraphics[width=1.4in]{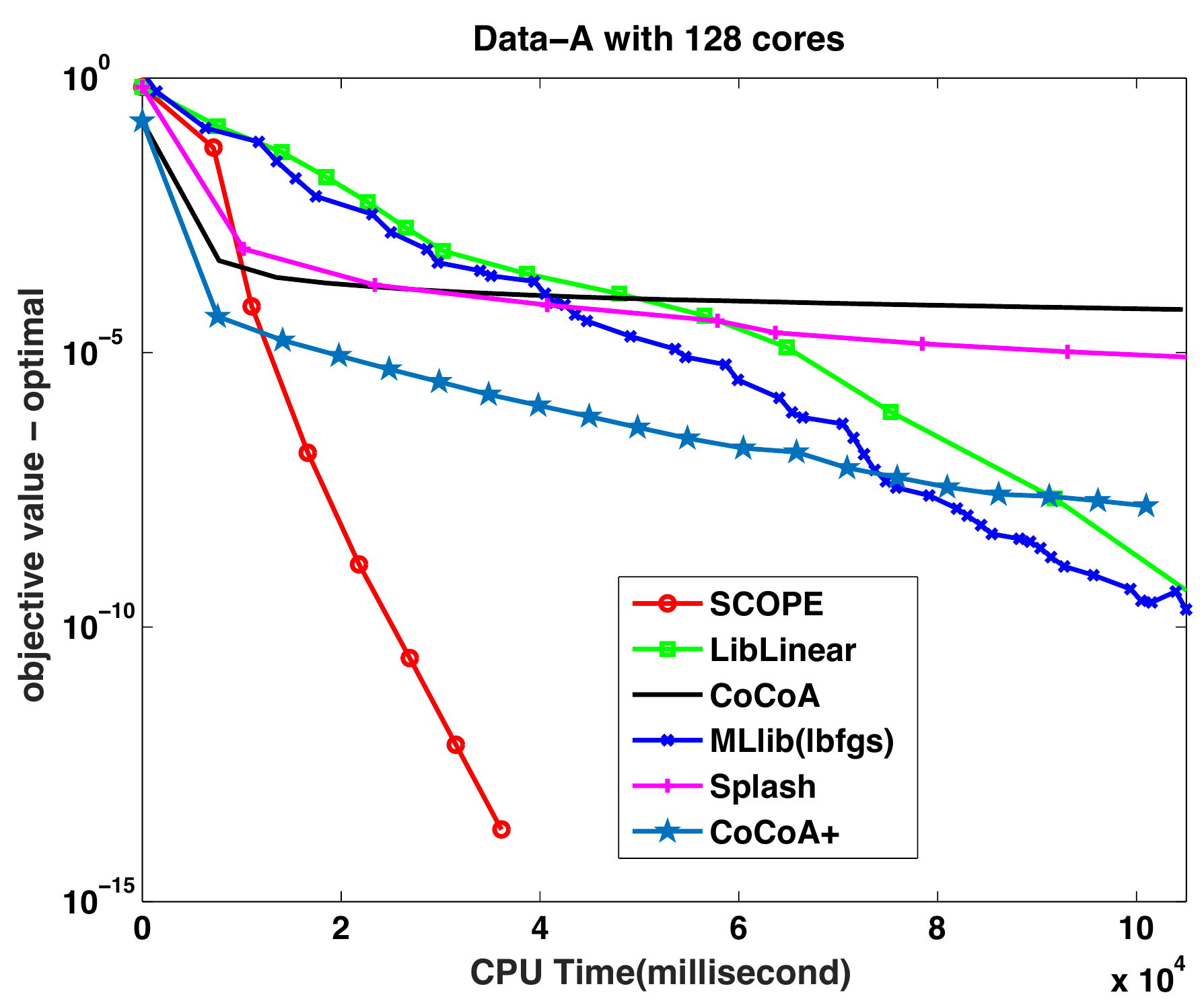}}
  \caption{Efficiency comparison with baselines.}\label{loss}
\end{figure}

We compare SCOPE with other baselines on the four datasets. The result is shown in Figure~\ref{loss}. Each marked point on the curves denotes one update for $\w$ by the Master, which typically corresponds to an iteration in the outer-loop. For SCOPE, good convergence results can be got with number of updates~(i.e., the $T$ in Algorithm~\ref{alg:mScope}) less than five. We can find that Splash vibrates on some datasets since it introduces variance in the training process. On the contrary, SCOPE are stable, which means that SCOPE is a variance reduction method like SVRG. It is easy to see that SCOPE has a linear convergence rate, which also conforms to our theoretical analysis. Furthermore, SCOPE is much faster than all the other baselines.

SCOPE can also outperform SVRGfoR~\cite{DBLP:journals/corr/KonecnyMR15} and DisSVRG. Experimental comparison can be found in appendix~\cite{DBLP:journals/corr/ZhaoXSGL16}.

\begin{figure}[!htb]
  \centering
  \includegraphics[width=1.3in]{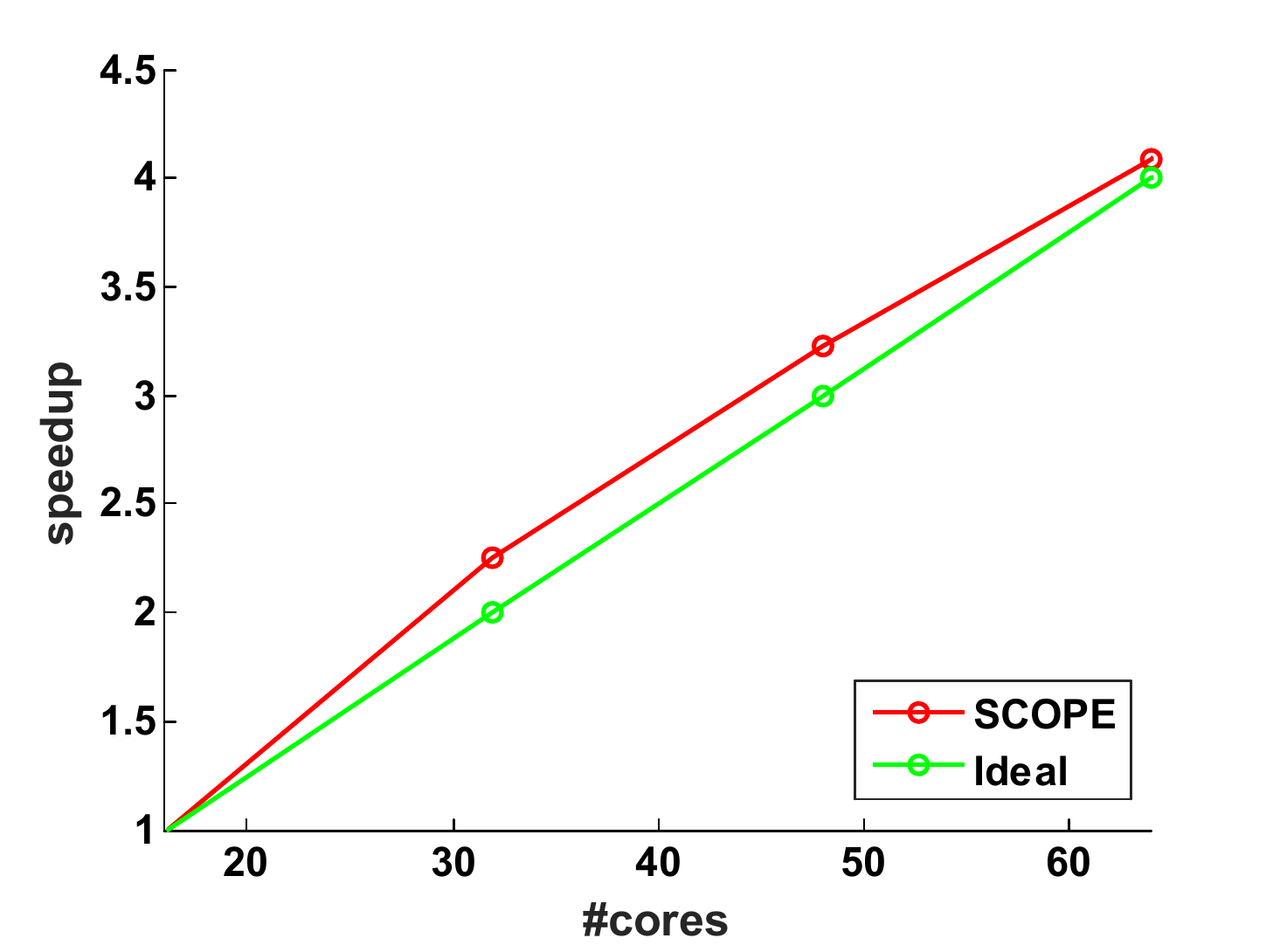}
  \caption{Speedup}\label{fig:SpeedupAndSynCost}
\end{figure}

\subsection{Speedup}
We use dataset MNIST-8M for speedup evaluation of SCOPE. Two cores are used for each machine. We evaluate speedup by increasing the number of machines. The training process will stop when the gap between the objective function value and the optimal value is less than $10^{-10}$. The $speedup$ is defined as follows:
$
speedup = \frac{time~with~16~cores~by~SCOPE}{time~with~2\pi~cores}
$
where $\pi$ is the number of machines and we choose $\pi = 8,16,24,32$. The experiments are performed by 5 times and the average time is reported for the final speedup result.

The speedup result is shown in Figure~\ref{fig:SpeedupAndSynCost}, where we can find that SCOPE has a super-linear speedup. This might be reasonable due to the higher cache hit ratio with more machines~\cite{DBLP:conf/cikm/YuSYL14}. This speedup result is quite promising on our multi-machine settings since the communication cost is much larger than that of multi-thread setting. The good speedup of SCOPE can be explained by the fact that most training work can be locally completed by each Worker and SCOPE does not need much communication cost.

SCOPE is based on the synchronous MapReduce framework of Spark. One shortcoming of synchronous framework is the synchronization cost, which includes both communication time and waiting time. We also do experiments to show the low synchronization cost of SCOPE, which can be found in the appendix~\cite{DBLP:journals/corr/ZhaoXSGL16}.

\section{Conclusion}
In this paper, we propose a novel DSO method, called SCOPE, for distributed machine
learning on Spark. Theoretical analysis shows that SCOPE is convergent with linear convergence rate for strongly convex
cases. Empirical results show that SCOPE can outperform other state-of-the-art distributed methods on Spark.




\section{Acknowledgements}
{This work is partially supported by the ``DengFeng" project of Nanjing University.}

\small
\bibliographystyle{aaai}
\bibliography{ref}

\newpage
\appendix

\section{Appendix}
\subsection{SVRG and Mini-Batch based Distributed SVRG}\label{sec:appendixSVRG}
The sequential SVRG is outlined in Algorithm~\ref{alg:seqSVRG}, which is the same as the original SVRG in~\cite{DBLP:conf/nips/Johnson013}.

\begin{algorithm}[ht]
\caption{Sequential SVRG}
\label{alg:seqSVRG}
\begin{algorithmic}
\STATE Initialization: initialize $\w_0, \eta$;
\FOR{$t=0,1,2,...,T$}
\STATE $\u_0 = \w_t$;
\STATE Compute the full gradient $\z = \frac{1}{n} \sum_{i=1}^n \nabla f_i(\u_0)$;
\FOR{$m=0$ to $M-1$}
\STATE Randomly pick up an $i_m$ from $\left\{ 1,\ldots,n \right\}$;
\STATE $\u_{m+1} = \u_{m} - \eta (\nabla f_{i_m}(\u_m) - \nabla f_{i_m}(\u_0) + \z)$;
\ENDFOR
\STATE Take $\w_{t+1}$ to be $\u_M$ or the average of $\{\u_m\}$;
\ENDFOR
\end{algorithmic}
\end{algorithm}


The mini-batch based distributed SVRG~(called DisSVRG) is outlined in Algorithm~\ref{alg:mSVRG} and Algorithm~\ref{alg:wSVRG}, with Algorithm~\ref{alg:mSVRG} for the operations completed by the Master and Algorithm~\ref{alg:wSVRG} for the operations completed by the Workers.

\begin{algorithm}[htb]
\caption{Task of Master in DisSVRG}
\label{alg:mSVRG}
\begin{algorithmic}
\STATE Initialization: $p$ Workers, $\w_0$, $\eta$;
\FOR{$t=0,1,2, \ldots,T$}
\STATE $\u_0 = \w_t$;
\STATE Send $\w_t$ to the Workers;
\STATE Wait until it receives $\z_1, \z_2, \ldots, \z_p$ from the $p$ Workers;
\STATE Compute the \emph{full gradient} $\z = \frac{1}{n} \sum_{k=1}^p \z_k$;
\FOR{$m=0$ to $M-1$}
\STATE Send $\u_m$ to all Workers;
\STATE Wait until it receives $|\SM_{m,k}|$, $\{\nabla f_{\SM_{m,k}}(\u_m)\}$ and $\{\nabla f_{\SM_{m,k}}(\w_t)\}$ from all the Workers, compute $|\SM_m| = \underset{k}{\sum} |\SM_{m,k}|$;
\STATE Compute $\nabla f_{\SM_m}(\u_m) = \frac{1}{|\SM_m|}\sum_{k=1}^p \nabla f_{\SM_{m,k}}(\u_m)$;
\STATE Compute $\nabla f_{\SM_m}(\u_0) = \frac{1}{|\SM_m|}\sum_{k=1}^p \nabla f_{\SM_{m,k}}(\w_t)$;
\STATE Update the parameter: $\u_{m+1} = \u_{m} - \eta (\nabla f_{\SM_m}(\u_m) - \nabla f_{\SM_m}(\u_0) + \z)$;
\ENDFOR
\STATE Take $\w_{t+1}$ to be $\u_M$ or the average of $\{\u_m\}$;
\ENDFOR
\end{algorithmic}
\end{algorithm}

\begin{algorithm}[htb]
\caption{Task of Workers in DisSVRG}
\label{alg:wSVRG}
\begin{algorithmic}
\STATE For the Worker$\_k$:
\FOR{$t=0,1,2, \ldots,T$}
\STATE Wait until it gets the newest parameter $\w_t$ from the Master;
\STATE Compute the \emph{local gradient sum} $\z_k = \sum_{i \in \DM_k} \nabla f_i(\w_t)$, and then send $\z_k$ to the Master;
\FOR{$m=0$ to $M-1$}
\STATE Wait until it gets the newest parameter $\u_m$ from the Master;
\STATE Randomly pick up a mini-batch indices $\SM_{m,k}$ from $\DM_k$;
\STATE Compute $\nabla f_{\SM_{m,k}}(\u_m) = \sum_{i \in \SM_{m,k}}\nabla f_i (\u_m)$;
\STATE Compute $\nabla f_{\SM_{m,k}}(\w_t) = \sum_{i \in \SM_{m,k}}\nabla f_i (\w_t)$;
\STATE Send $|\SM_{m,k}|$, $\nabla f_{\SM_{m,k}}(\u_m)$ and $\nabla f_{\SM_{m,k}}(\w_t)$ to the Master;
\ENDFOR
\ENDFOR
\end{algorithmic}
\end{algorithm}

%

\subsection{Proof of Lemma~\ref{lemma:gamma_m}}
We define the local stochastic gradient in Algorithm~\ref{alg:wScope} as follows:
\begin{align}
\v_{k,m} = \nabla f_{i_{k,m}}(\u_{k,m}) - \nabla f_{i_{k,m}}(\w_t) + \z + c(\u_{k,m} - \w_t). \nonumber
\end{align}
Then the update rule at local Workers can be rewritten as follows:
\begin{align}\label{update}
\u_{k,m+1} = \u_{k,m} - \eta \v_{k,m}.
\end{align}

First, we give the expectation and variance property of $\v_{k,m}$ in Lemma~\ref{lemma:Ev} and Lemma~\ref{lemma:Var}.

\begin{lemma}\label{lemma:Ev}
The conditional expectation of local stochastic gradient $\v_{k,m}$ on $\u_{k,m}$ is
\begin{align}
   &\EB[ \v_{k,m} | \u_{k,m} ] \nonumber \\
= &\nabla F_k(\u_{k,m}) - \nabla F_k(\w_t) + \z + c(\u_{k,m} - \w_t). \nonumber
\end{align}
\end{lemma}

\begin{proof}
\begin{align}
   &\EB[ \v_{k,m} | \u_{k,m} ] \nonumber \\
= &\frac{1}{q} \sum_{i \in \DM_k} [\nabla f_i(\u_{k,m}) - \nabla f_i(\w_t) + \z + c(\u_{k,m} - \w_t)] \nonumber \\
= &\nabla F_k(\u_{k,m}) - \nabla F_k(\w_t) + \z + c(\u_{k,m} - \w_t) \nonumber
\end{align}
\end{proof}

\begin{lemma}\label{lemma:Var}
The variance of $\v_{k,m}$ has the the following property:
\begin{align}
       &\EB[ \| \v_{k,m} \|^2 | \u_{k,m} ]  \nonumber \\
& \leq 3(L^2 + c^2) \| \u_{k,m} - \w_t \|^2 + 3L^2\| \w_t - \w^* \|^2. \nonumber
\end{align}
\end{lemma}
\begin{proof}
\begin{align}
       &\EB[ \| \v_{k,m} \|^2 | \u_{k,m} ] \nonumber \\
=     &\frac{1}{q} \sum_{i \in \DM_k} \| \nabla f_i(\u_{k,m}) - \nabla f_i(\w_t) + \z + c(\u_{k,m} - \w_t) \|^2 \nonumber \\
\leq &\frac{3}{q} \sum_{i \in \DM_k} [\| \nabla f_i(\u_{k,m}) - \nabla f_i(\w_t) \|^2 + \| \z \|^2 \nonumber \\
& + c^2\| \u_{k,m} - \w_t \|^2] \nonumber \\
\leq &\frac{3}{q} \sum_{i \in \DM_k} \left[ L^2 \| \u_{k,m} - \w_t \|^2 + \| \z \|^2 + c^2\| \u_{k,m} - \w_t \|^2\right] \nonumber \\
\leq &\frac{3}{q} \sum_{i \in \DM_k} \left[ (L^2 + c^2) \| \u_{k,m} - \w_t \|^2 + L^2\| \w_t - \w^* \|^2 \right] \nonumber \\
=     &3(L^2 + c^2) \| \u_{k,m} - \w_t \|^2 + 3L^2\| \w_t - \w^* \|^2 \nonumber
\end{align}
The second inequality uses Assumption \ref{ass:smooth}. The third inequality uses the fact that $\nabla P(\w^*) = 0$.
\end{proof}

%

Based on Lemma~\ref{lemma:Ev} and Lemma~\ref{lemma:Var}, we prove Lemma~\ref{lemma:gamma_m} as follows:
\begin{proof}
According to (\ref{update}), we have
\begin{align}
     &\| \u_{k,m+1} - \w^* \|^2 \nonumber \\
=   &\| \u_{k,m} - \w^* - \eta \v_{k,m} \|^2 \nonumber \\
=   &\| \u_{k,m} - \w^* \|^2 - 2\eta \v_{k,m}^T(\u_{k,m} - \w^*) + \eta^2\| \v_{k,m} \|^2 \nonumber
\end{align}

We take expectation on both sides of the above equality, and obtain
\begin{align}\label{the:eq1}
\EB [\| \u_{k,m+1} &- \w^* \|^2 | \u_{k,m} ] = \| \u_{k,m} - \w^* \|^2 \nonumber \\
     &- 2\eta (\nabla F_k(\u_{k,m}) - \nabla F_k(\w_t) + \z \nonumber \\
     &+ c(\u_{k,m} - \w_t))^T(\u_{k,m} - \w^*) \nonumber \\
     &+\eta^2 \EB[\| \v_{k,m} \|^2 | \u_{k,m}]
\end{align}

For the second line of the right side of the above equality, we have
\begin{align}
        &\EB[\v_{k,m}|\u_{k,m}]^T(\u_{k,m} - \w^*) \nonumber \\
=       &\nabla F_k(\u_{k,m})^T(\u_{k,m} - \w^*) \nonumber \\
        &+ (\z - \nabla F_k(\w_t))^T(\u_{k,m} - \w^*) \nonumber \\
        &+ c(\u_{k,m} - \w_t)^T(\u_{k,m} - \w^*)  \nonumber \\
\geq    &F_k(\u_{k,m}) - F_k(\w^*) + \frac{\mu}{2} \| \u_{k,m} - \w^* \|^2 \nonumber \\
        &+\z^T(\u_{k,m} - \w_t) - \nabla F_k(\w_t)^T(\u_{k,m} - \w_t) \nonumber \\
        &+(\z - \nabla F_k(\w_t))^T(\w_t - \w^*) \nonumber \\
        &+c(\u_{k,m} - \w_t)^T(\u_{k,m} - \w^*)  \nonumber \\
\geq    &F_k(\u_{k,m}) - F(\w^*) + \frac{\mu}{2} \| \u_{k,m} - \w^* \|^2 \nonumber \\
        &+F_k(\w_t) - F_k(\u_{k,m}) + \frac{\mu}{2} \| \u_{k,m} - \w_t \|^2 \nonumber \\
        &+\z^T(\u_{k,m} - \w_t) + (\z - \nabla F_k(\w_t))^T(\w_t - \w^*) \nonumber \\
        &+c(\u_{k,m} - \w_t)^T(\u_{k,m} - \w^*)  \nonumber \\
=       &F_k(\w_t) - F_k(\w^*) \nonumber \\
        &+ \frac{\mu}{2} \| \u_{k,m} - \w^* \|^2 + \frac{\mu}{2} \| \u_{k,m} - \w_t \|^2 \nonumber \\
        &+\z^T(\u_{k,m} - \w_t) + (\z - \nabla F_k(\w_t))^T(\w_t - \w^*) \nonumber \\
        &+ c(\u_{k,m} - \w_t)^T(\u_{k,m} - \w^*)  \nonumber \\
=       &F_k(\w_t) - F_k(\w^*) \nonumber \\
        &+ \frac{\mu+c}{2} \| \u_{k,m} - \w^* \|^2 + \frac{\mu+c}{2} \| \u_{k,m} - \w_t \|^2 \nonumber \\
        &+\z^T(\u_{k,m} - \w_t) + (\z - \nabla F_k(\w_t))^T(\w_t - \w^*) \nonumber \\
        &-\frac{c}{2} \{ \| \u_{k,m} - \w^* \|^2 - 2(\u_{k,m} - \w_t)^T(\u_{k,m} - \w^*) \nonumber \\
        &+\| \u_{k,m} - \w_t \|^2 \} \nonumber \\
=       &F_k(\w_t) - F_k(\w^*) \nonumber \\
        &+ \frac{\mu+c}{2} \| \u_{k,m} - \w^* \|^2 + \frac{\mu+c}{2} \| \u_{k,m} - \w_t \|^2 \nonumber \\
        &+\z^T(\u_{k,m} - \w_t) + (\z - \nabla F_k(\w_t))^T(\w_t - \w^*) \nonumber \\
        &-\frac{c}{2} \| \w_t - \w^* \|^2 \nonumber
\end{align}
Both the first and second inequalities for the above derivation use Assumption~\ref{ass:stroconv}.

We use $\sigma_m = \sigma(\u_{1,m}, \u_{2,m}, \ldots, \u_{p,m})$ to denote the $\sigma$-algebra. Then we can take a summation for (\ref{the:eq1}) with $k=1$ to $p$, and obtain
\begin{align}\label{the:ineq1}
       &\sum_{k=1}^p \EB [\| \u_{k,m+1} - \w^* \|^2 | \sigma_m ] \nonumber \\
\leq   &\sum_{k=1}^p  \| \u_{k,m} - \w^* \|^2 \nonumber \\
       &-2\eta \sum_{k=1}^p \{F_k(\w_t) - F_k(\w^*) + \frac{\mu+c}{2} \| \u_{k,m} - \w^* \|^2 \nonumber \\
       &+\frac{\mu+c}{2} \| \u_{k,m} - \w_t \|^2 + \z^T(\u_{k,m} - \w_t) \nonumber \\
       &-\frac{c}{2} \| \w_t - \w^* \|^2 + (\z - \nabla F_k(\w_t))^T(\w_t - \w^*)\}  \nonumber \\
       &+\eta^2 \sum_{k=1}^p \EB[\| \v_{k,m} \|^2 | \sigma_m] \nonumber \\
=      &\sum_{k=1}^p  \| \u_{k,m} - \w^* \|^2 \nonumber \\
       &-2\eta \sum_{k=1}^p \{P(\w_t) - P(\w^*) + \frac{\mu+c}{2} \| \u_{k,m} - \w^* \|^2 \nonumber \\
       &+\frac{\mu+c}{2} \| \u_{k,m} - \w_t \|^2 + \z^T(\u_{k,m} - \w_t)\} \nonumber \\
       &+cp\eta \| \w_t - \w^* \|^2 + \eta^2 \sum_{k=1}^p \EB[\| \v_{k,m} \|^2 | \sigma_m] \nonumber \\
\leq   &\sum_{k=1}^p  \| \u_{k,m} - \w^* \|^2 \nonumber \\
       &-2\eta \sum_{k=1}^p \{P(\w_t) - P(\w^*) + \frac{\mu+c}{2} \| \u_{k,m} - \w^* \|^2 \nonumber \\
       &+\frac{\mu+c-L}{2} \| \u_{k,m} - \w_t \|^2 + P(\u_{k,m}) - P(\w_t)\} \nonumber \\
       &+cp\eta \| \w_t - \w^* \|^2 + \eta^2 \sum_{k=1}^p \EB[\| \v_{k,m} \|^2 | \sigma_m] \nonumber \\
=      &\sum_{k=1}^p  \| \u_{k,m} - \w^* \|^2 \nonumber \\
       &-2\eta \sum_{k=1}^p \{P(\u_{k,m}) - P(\w^*) + \frac{\mu+c}{2} \| \u_{k,m} - \w^* \|^2 \nonumber \\
       &+\frac{\mu+c-L}{2} \| \u_{k,m} - \w_t \|^2 \} \nonumber \\
       &+cp\eta \| \w_t - \w^* \|^2 + \eta^2 \sum_{k=1}^p \EB[\| \v_{k,m} \|^2 | \sigma_m] \nonumber \\
\leq   &\sum_{k=1}^p  \| \u_{k,m} - \w^* \|^2 -2\eta \sum_{k=1}^p \{\frac{2\mu+c}{2} \| \u_{k,m} \nonumber \\
&- \w^* \|^2 +\frac{\mu+c-L}{2} \| \u_{k,m} - \w_t \|^2 \} \nonumber \\
       &+cp\eta \| \w_t - \w^* \|^2 + \eta^2 \sum_{k=1}^p \EB[\| \v_{k,m} \|^2 | \sigma_m]
\end{align}
In the first equality, we use the definition of local function $F_k(\cdot)$ that $P(\w) = \frac{1}{p}\sum_{k=1}^p \nabla F_k(\w)$ which leads to $\sum_{k=1}^p F_k(\w_t) = \sum_{k=1}^p P(\w_t)$ and $\sum_{k=1}^p (\z - \nabla F_k(\w_t))^T(\w_t - \w^*) = 0$. In the first inequality, we use Assumption \ref{ass:smooth} which leads to $P(\u_{k,m}) \leq P(\w_t) + \z^T(\u_{k,m} - \w_t) + \frac{L}{2}\| \u_{k,m} - \w_t \|^2$.

If we use $\gamma_m = \frac{1}{p} \sum_{k=1}^p \EB \| \u_{k,m} - \w^* \|^2$, then according to Algorithm \ref{alg:wScope}, it is easy to get that $\gamma_0 = \EB \| \w_t - \w^* \|^2$. Moreover, according to (\ref{the:ineq1}) and Lemma \ref{lemma:Var}, we can obtain
\begin{align}
\gamma_{m+1} \leq \gamma_m - \eta(2\mu + c)\gamma_m + (c\eta + 3L^2 \eta^2)x_0 + a_m \nonumber
\end{align}
where
\begin{align}\label{def:am}
a_m = \frac{3\eta^2(L^2 + c^2) - \eta(\mu + c - L)}{p} \sum_{k=1}^p\EB \| \u_{k,m} - \w_t \|^2 \nonumber
\end{align}
If $c > L - \mu$, we can choose a small $\eta$ such that $a_m < 0$. Then we get the result
\begin{align}
\gamma_{m+1} \leq [1 - \eta(2\mu + c)]\gamma_m + (c\eta + 3L^2 \eta^2)\gamma_0 \nonumber
\end{align}
\end{proof}

\subsection{Proof of Theorem~\ref{cor1}}
\begin{proof}
According to Lemma~\ref{lemma:gamma_m}, we have
\begin{align}
\gamma_{m} &\leq \alpha \gamma_{m-1} + \beta \gamma_0 \nonumber \\
          &\leq (\alpha^m + \frac{\beta}{1 - \alpha})\gamma_0 \nonumber
\end{align}
Since we take $\w_{t+1} = \frac{1}{p} \sum_{k=1}^p \u_{k,M}$, then we have
\begin{align}
\EB \| \w_{t+1} - \w^* \|^2 =    &\EB \| \frac{1}{p} \sum_{k=1}^p \u_{k,M} - \w^* \|^2 \nonumber \\
                                        \leq &\frac{1}{p} \sum_{k=1}^p \EB \| \u_{k,M} - \w^* \|^2 \nonumber \\
                                        =     &\gamma_M \nonumber \\
                                        \leq &(\alpha^M + \frac{\beta}{1 - \alpha})\EB \| \w_t - \w^* \|^2 \nonumber
\end{align}
\end{proof}

\subsection{Proof of Theorem~\ref{cor2}}
\begin{proof}
According to Lemma~\ref{lemma:gamma_m}, we have
\begin{align}
\gamma_{m+1} + (1-\alpha)\gamma_m \leq \gamma_{m} + \beta \gamma_0 \nonumber
\end{align}
Summing $m$ from $1$ to $M$, we have
\begin{align}
\gamma_{m+1} + (1-\alpha)\sum_{m=1}^{M} \gamma_m \leq (1 + M\beta)\gamma_0 \nonumber
\end{align}
Since we take $\w_{t+1} = \frac{1}{pM} \sum_{m=1}^{M} \sum_{k=1}^p \u_{k,m}$, then we have
\begin{align}
\EB \| \w_{t+1} - \w^* \|^2 =    &\EB \| \frac{1}{pM} \sum_{m=1}^{M} \sum_{k=1}^p \u_{k,M} - \w^* \|^2 \nonumber \\
                                        \leq &\frac{1}{pM} \sum_{m=1}^{M} \sum_{k=1}^p \| \u_{k,M} - \w^* \|^2 \nonumber \\
                                        \leq &(\frac{1}{M(1-\alpha)} + \frac{\beta}{1-\alpha})\EB \| \w_t - \w^* \|^2 \nonumber
\end{align}
\end{proof}

\subsection{Efficiency Comparison with DisSVRG}
\begin{figure}[htb]
  \centering
  \includegraphics[width=2.8in]{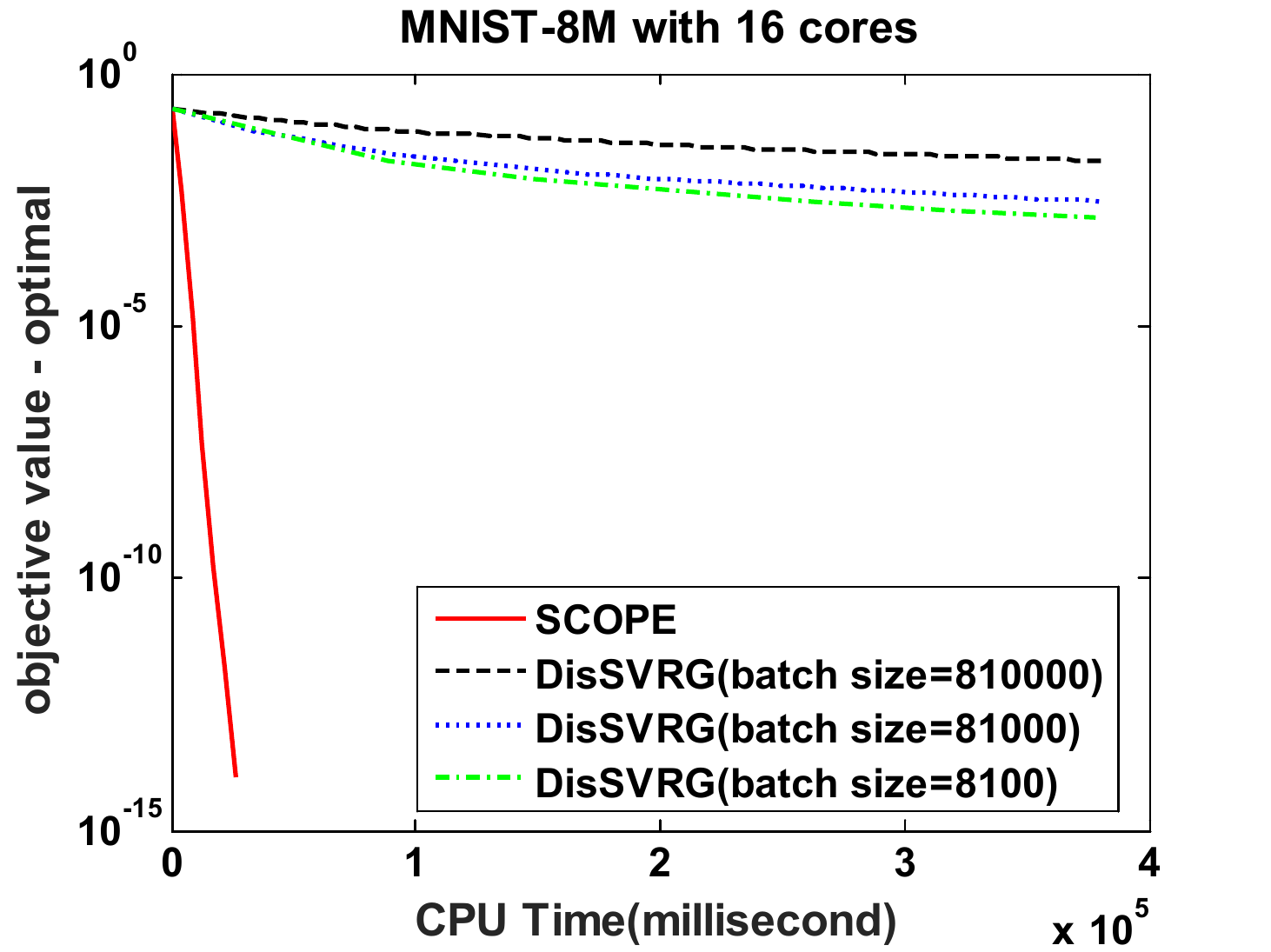}
 \vspace{-0.2cm}
  \caption{Efficiency comparison with DisSVRG}\label{dissvrg}
\end{figure}

In this section, we compare SCOPE with the mini-batch based distributed SVRG~(called DisSVRG) in Algorithm~\ref{alg:mSVRG} and Algorithm~\ref{alg:wSVRG} using the dataset MNIST-8M. The result is shown in Figure~\ref{dissvrg}. The x-axis is the CPU time, containing both computation and synchronization time, with the unit millisecond. The y-axis is the objective function value minus the optimal value in a log scale. In this paper, the optimal value is the minimal value got by running all the baselines and SCOPE for a large number of iterations. It is easy to see that DisSVRG is much slower than our SCOPE, which means that the traditional mini-batch based DSO strategy is not scalable due to huge communication cost.
\vspace{-0.3cm}

\subsection{Efficiency Comparison with SVRGfoR}

\begin{figure}[htb]
  \centering
  \includegraphics[width=2.8in]{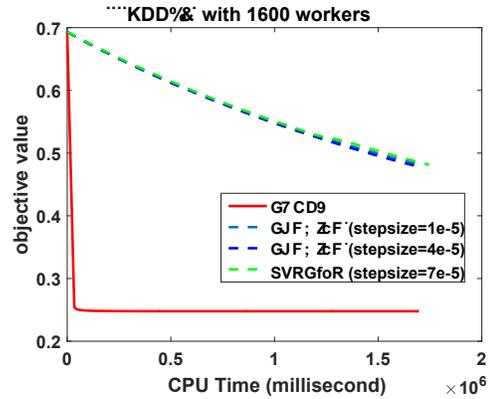}
 \vspace{-0.2cm}
  \caption{Efficiency comparison with SVRGfoR}\label{fig:SVRGfoR}
\end{figure}

SVRGfoR~\cite{DBLP:journals/corr/KonecnyMR15} is proposed for cases when the number of Workers is relatively large, and with unbalanced data partitions and sparse features. We use the KDD12 dataset with sparse features for evaluation. We construct a large cluster with 1600 Workers. Furthermore, we partition the data in an unbalanced way. The largest number of data points on one Worker is 423954, and the smallest number of data points on one Worker is 28. We tune several stepsizes for SVRGfoR to get the best performance. The experimental results are shown in Figure~\ref{fig:SVRGfoR}. We can find that SCOPE is much faster than SVRGfoR.

\subsection{Synchronization Cost}
SCOPE is based on the synchronous MapReduce framework of Spark. One shortcoming of synchronous framework is that there exists synchronization cost besides the computation cost. The synchronization cost includes both communication time and waiting time. Fortunately, the synchronization cost of SCOPE is low because most computation is completed locally and only a small number of synchronization times is needed. Here, we use experiment to verify this.

\begin{figure}[!htb]
  \centering
\includegraphics[width=2in]{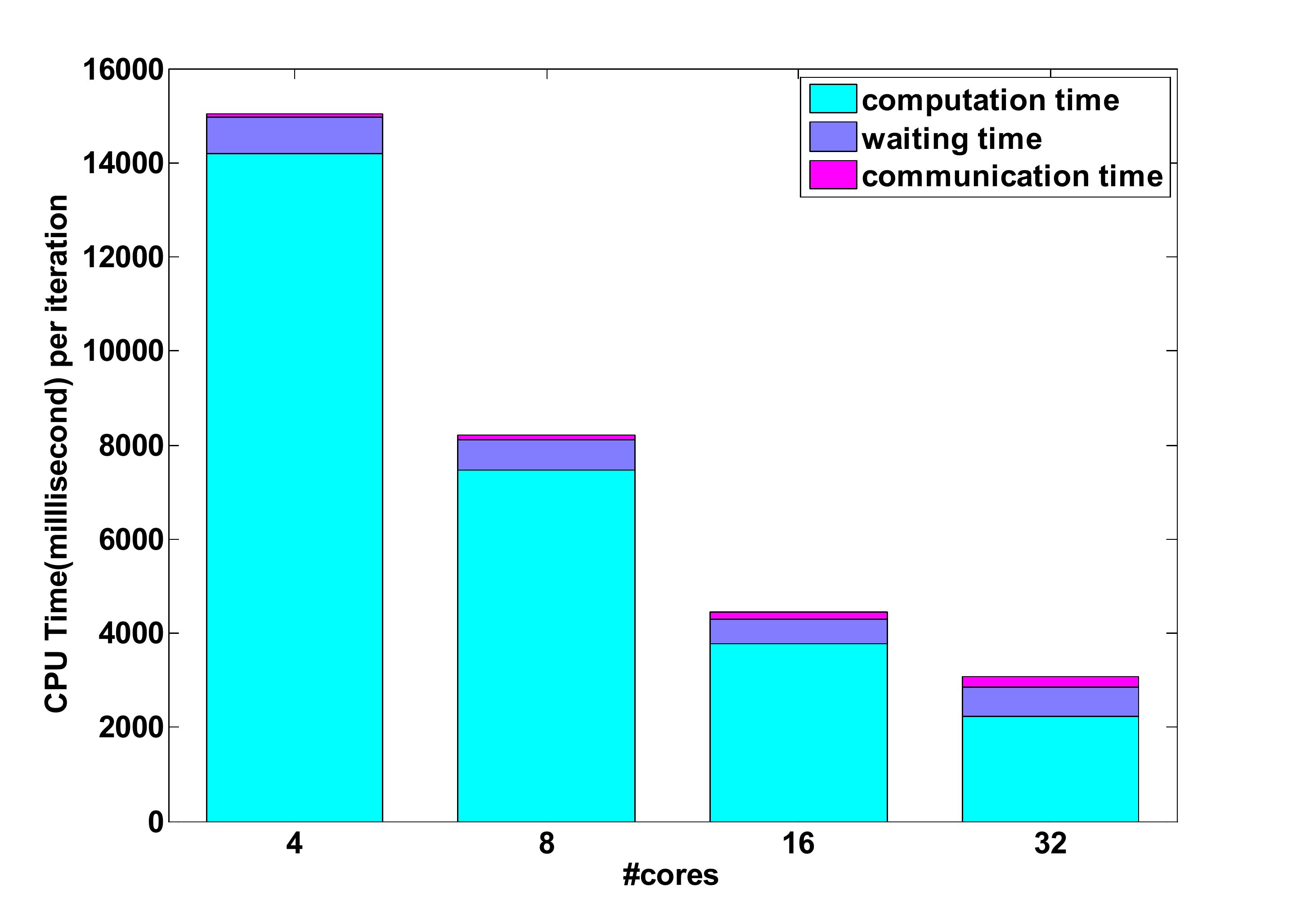}
  \vspace{-0.3cm}
  \caption{Synchronization cost.}\label{fig:SynCost}
\end{figure}

We use the dataset MNIST-8M for evaluation. The result is shown in Figure~\ref{fig:SynCost}. The x-axis is the number of cores, the y-axis is the CPU time~(in millisecond) per iteration, which is computed as dividing the total time by the number of iterations $T$. Please note that the CPU time includes both computation time and synchronization time~(cost). During the training process, if the Workers or Master are computing, we consider the time as computation time. In each synchronization step, we consider the time gap between the completion of the fastest Worker and the slowest Worker as waiting time. If there is communication between Workers and Master, we consider the time as communication time. From Figure~\ref{fig:SynCost}, we can find that SCOPE does not take too much waiting time and communication time compared to the computation time. We can also find that with the increase of the number of Workers~(cores), the synchronization time per iteration does not increase too much, which is promising for distributed learning on clusters with multiple machines.

The speedup in Figure~\ref{fig:SynCost} seems to be smaller than that in Figure~\ref{fig:SpeedupAndSynCost}. Both Figure~\ref{fig:SynCost} and Figure~\ref{fig:SpeedupAndSynCost} are correct. Some Workers still perform computation during the waiting time. So there is a repeating part in the waiting time and computation time in Figure~\ref{fig:SynCost}. Furthermore, the total number of iterations to achieve the same objective value may not be the same for different number of cores.

\end{document}